\documentclass[twoside]{article}

\usepackage[accepted]{aistats2020}

\usepackage{amsmath,amsfonts,bm}

\def\eqref#1{equation~\ref{#1}}

\def\1{\bm{1}}

\def\vmu{{\bm{\mu}}}
\def\vnu{{\bm{\nu}}}
\def\vtheta{{\bm{\theta}}}
\def\vvartheta{{\bm{\vartheta}}}
\def\vphi{{\bm{\phi}}}

\def\vx{{\bm{x}}}
\def\vy{{\bm{y}}}

\def\mW{{\bm{W}}}

\DeclareMathAlphabet{\mathsfit}{\encodingdefault}{\sfdefault}{m}{sl}
\SetMathAlphabet{\mathsfit}{bold}{\encodingdefault}{\sfdefault}{bx}{n}

\def\gT{{\mathcal{T}}}

\def\sR{{\mathbb{R}}}

\newcommand{\sigmoid}{\sigma}

\newcommand{\bert}{\mathrm{BERT}}
\newcommand{\attn}{\mathcal{A}}
\newcommand{\lnorm}{\mathcal{LN}}
\newcommand{\dout}{\mathcal{DO}}
\newcommand{\gelu}{\mathrm{GeLU}}
\newcommand{\bern}{\mathrm{Bern}}

\makeatletter
\renewcommand{\paragraph}{%
  \@startsection{paragraph}{4}%
  {\z@}{0.25ex \@plus 0.25ex \@minus .5ex}{-1em}%
  {\normalfont\normalsize\bfseries}%
}
\makeatother

\usepackage{hyperref}
\usepackage{url}

\usepackage{xcolor}
\usepackage{times}
\usepackage{epsfig}
\usepackage{graphicx}
\usepackage{amsmath}
\usepackage{amssymb}
\usepackage{booktabs}
\usepackage{multirow}
\usepackage[labelfont=bf,justification=justified,format=plain]{caption}
\usepackage{subcaption}
\usepackage[title]{appendix}
\usepackage[round]{natbib}

\begin{document}

\twocolumn[

\aistatstitle{How fine can fine-tuning be?\ \ Learning efficient language models}

\aistatsauthor{ 
    Evani Radiya-Dixit 
    \And 
    Xin Wang
}

\aistatsaddress{ 
    Stanford University\footnotemark
    \And 
    Cerebras Systems\footnotemark
} 
]
\addtocounter{footnote}{-1}
\footnotetext{Work done during internship at Cerebras Systems (current email: \texttt{\href{mailto:evanir@stanford.edu}evanir@stanford.edu}).}
\addtocounter{footnote}{1}
\footnotetext{Correspondence to XW (\texttt{\href{mailto:poincare.disk@gmail.com}poincare.disk@gmail.com}).}

\begin{abstract}

State-of-the-art performance on language understanding tasks is now achieved with increasingly large networks; the current record holder has billions of parameters.  
Given a language model pre-trained on massive unlabeled text corpora, only very light supervised fine-tuning is needed to learn a task: the number of fine-tuning steps is typically five orders of magnitude lower than the total parameter count.  
Does this mean that fine-tuning only introduces \emph{small} differences from the pre-trained model in the parameter space?  
If so, can one avoid storing and computing an entire model for each task?  
In this work, we address these questions by using Bidirectional Encoder Representations from Transformers (BERT) as an example.  
As expected, we find that the fine-tuned models are close in parameter space to the pre-trained one, with the closeness varying from layer to layer.  
We show that it suffices to fine-tune only the most critical layers.  
Further, we find that there are surprisingly many \emph{good} solutions in the set of sparsified versions of the pre-trained model.  
As a result, fine-tuning of huge language models can be achieved by simply setting a certain number of entries in certain layers of the pre-trained parameters to zero, saving both task-specific parameter storage and computational cost. 

\end{abstract}

\section{Introduction}
\label{sec:intro}

Modern deep neural networks operate in a regime where the generalization gap diminishes with growing model capacity, defying the classical bias-variance tradeoff~\citep{Belkin2018}.  
Increased model capacity always leads to better generalization, a trend highly prominent in the natural language understanding domain.  
From BERT~\citep[340M parameters,][]{Devlin2018}, to GPT-2~\citep[1.5B parameters,][]{Radford2018}, and to Megatron-LM~\citep[8.3B parameters,][]{Shoeybi2019}, state-of-the-art performance in language comprehension tasks keeps improving with larger model capacity.  

These huge language models are first pre-trained on large text corpora.  
\emph{Pre-training} is a learning procedure, often unsupervised, that yields a good common initialization for further supervised learning of various downstream tasks.  This further learning, called \emph{fine-tuning}, is an additional optimization of model parameters jointly with a very small number of extra task-specific parameters (\textit{e.g.} Table~\ref{tab:param}).  

Though larger models generalize better, they are more expensive computationally, and the costs grow with the number of tasks learned.  
The high computational cost is usually addressed by network compression techniques that produce compact models for efficient inference~\citep[\textit{e.g.}][]{Zhao2019}. 
To reduce task-specific cost, transfer learning and continual learning methods are useful to maximize sharing of parameters across tasks~\citep[\textit{e.g.}][]{Houlsby2019,Liu2019}.  
In this work, we attempt to achieve these two desirable outcomes with a single effort.  
We use Bidirectional Encoder Representations from Transformers~\citep[BERT,][]{Devlin2018} and the General Language Understanding Evaluation (GLUE) benchmark~\citep{Wang2018} as a working example.  

Our intuition comes from the observation that the amount of fine-tuning necessary to learn each task is very small (five orders of magnitude smaller than the dimensionality of the parameter space, Table~\ref{tab:param}).  
This is not surprising: the high quality of a pre-trained model should naturally lead to rather few iterations needed to fine-tune it to perform specific tasks. 
But importantly, such light fine-tuning might result in fine-tuned models hypothetically closer\,\footnote{
  This vague notion of \emph{closeness}, \textit{viz.} separation by \emph{few} gradient update steps in the parameter space, will be made explicit later in the text.  
} to the pre-trained one in parameter space.  
This suggests a potentially high degree of computational redundancy across tasks that might be avoided at inference time. 

\begin{figure}[t]
    \centering
    \includegraphics[width=0.48\textwidth]{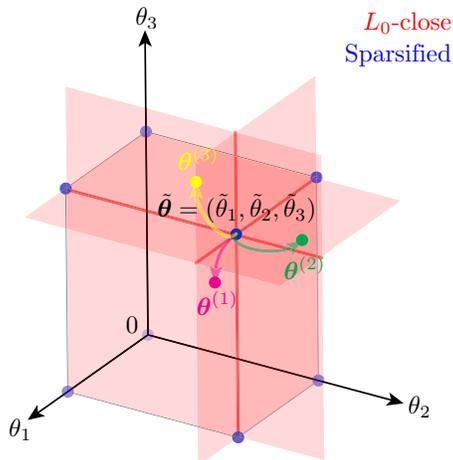}
    \caption{
    An illustration of $L_0$-close and sparsification constraints on a pre-trained parameter in a three-dimensional parameter space.
    Here $\tilde \vtheta$ is the pre-trained parameters.  
    Individual fine-tuning procedures for different tasks send the pre-trained parameters to distinct optimized parameters in a close $L_1$-neighborhood, \textit{e.g.} $\vtheta^{(1)}$, $\vtheta^{(2)}$ and $\vtheta^{(3)}$, of which each component, \textit{viz.}  $\theta_1$, $\theta_2$ or $\theta_3$, is subject to change.  
    The $L_0$-closeness constraint (red with its saturation encoding closeness) forces optimization within those parameter configurations that share a certain fraction of components with $\tilde \vtheta$, \textit{i.e.} having a small number of different components.  
    The sparsification constraint (blue with its saturation encoding density) further confines optimization to a discrete subset of $L_0$-close parameters, where all changed components have to be set to zero.  
    }
    \label{fig:cart}
\end{figure}

We first observe that the fine-tuned and pre-trained parameters are both $L_1$-close and angular-close in parameter space, consistent with the small number of fine-tuning iterations separating them.  
Despite this closeness in parameter space, the fine-tuned parameters are not constrained to share any components with the pre-trained weights, and thus are equally expensive to store and to compute per iteration. 
We conjecture that there also exist good fine-tuned parameters under efficient computational constraints, even though they might be more $L_1$-distant or angular-distant. 
In order to make fine-tuned models share parameters with the pre-trained models, we optimize parameters $L_0$-close to pre-trained (Figure~\ref{fig:cart}, red) by fine-tuning only the most sensitive layers (\textit{i.e.} those most distant in parameter subspaces) of the network.  
Furthermore, we attempt to learn a task by sparsifying the pre-trained weights (Figure~\ref{fig:cart}, blue).  
Surprisingly, our results reveal an abundance of good task-specific parameter configurations within sparified versions of pre-trained models: a specific task can be learned by simply masking anywhere between $1\%$ to $40\%$ of the pre-trained weights to zero.

A major contribution of the present work is the demonstration that fine-tuning can be realized by sparsification, which has favorable practical implications.
By forcing fine-tuned parameters to be $L_0$-close to the pre-trained ones, one only needs to store a small number of different weights per task, in addition to the common pre-trained weights, substantially saving parameter memory when there are many tasks to perform.
By forcing fine-tuned parameters to be sparse, one potentially saves both memory and compute, because each task only requires a binary mask on top of the common pre-trained parameters and sparse linear algebraic operations could be used instead of dense ones.

\section{Related work}
\label{sec:rel_work}

A large body of literature is concerned with sparsification of large networks for efficient inference~\citep[\textit{e.g.}][]{Zhu2017}.  
Our search for $L_0$-close fine-tuning solutions is motivated by the observation that sensitivities of the optimization objective to different layers in a network are highly variable~\citep{Zhang2019}.
\cite{Zhou2019} trained sparse connectivity patterns over randomly initialized network parameters, termed \emph{supermasks}, suggesting that sparsification plays a role similar and complementary to gradient-based learning of the objective. 
This is also related to network architecture search (NAS).

The most similar study to ours is \emph{piggyback} and its variants~\citep{Mallya2018,Mancini2019}, where in a multi-task visual object classification scenario, the authors trained task-specific binary masks on top of a shared set of pre-trained parameters. 
In this work, we not only applied similar techniques to larger pre-trained language models, but also studied the sparsity-accuracy tradeoff in a systematic way.  
\citet{Houlsby2019} added adapter modules to pre-trained language models, achieving parameter sharing across multiple tasks, but not reducing the computational cost of the resultant fine-tuned networks.  
Also, note that randomly generated high-dimensional masks can also support multi-task learning, \textit{e.g.} \citet{Cheung2019}.

To impose parameter sparseness differentiably in combination with the $L_0$-closeness constraint, instead of principled approaches to imposing $L_0$-regularization \citep{Louizos2017}, we used the simpler straight-through estimator, much like binary quantization techniques~\citep{Courbariaux2015,Courbariaux2016}; note that this is also used by \citet{Mallya2018} and \citet{Zhou2019}.  

\section{Methods}
\label{sec:meth}

\begin{table*}[t]
\caption{Task-specific model information of $\bert_\text{BASE}$ (parameter count 109M).}
\label{tab:param}
\vspace{-3mm}
\begin{center}
\begin{small}
\setlength\tabcolsep{3.6pt}
\begin{tabular}{l r r r r r r r r }
    \toprule
    {GLUE Task} & MNLI & QQP & QNLI & SST-2 & CoLA & STS-B & MRPC & RTE \\ 
    \midrule
    {Additional parameter count}     &    $2,304$  &  $1,536$ & $1,536$  & $1,536$ & $1,536$ & $768$ & $1,536$ & $1,536$ \\
    {Fine-tuning iteration count}    &   $36,816$  & $34,113$ & $9,822$  & $6,315$ &   $804$ & $540$ &   $345$ &   $234$ \\
    \bottomrule
\end{tabular}
\end{small}
\end{center}  
\vskip -0.1in
\label{tab:task_info}
\end{table*}

\subsection{Notations and model architecture}
Consider a pre-trained network $F_\vtheta: \vx \mapsto F(\vx; \vtheta)$ that transforms input sequence $\vx$ into a good representation. 
It is parameterized by $\vtheta$, noted as subscript for convenience.  
The fine-tuning procedure to perform a task $t \in \gT$ ($\gT$ being the set of tasks) can be described as a supervised training procedure of model $G^{(t)}_\vphi \circ F_\vtheta : \vx \mapsto \vy$ on fine-tuning set $\left\{ (\vx^{(t)}_i, \vy^{(t)}_i) \right\}$. $G^{(t)}_\vphi$ is a task-specific last layer unique to task $t$ and is parameterized by $\vphi$; $\circ$ denotes function composition. 

In the case of BERT, we have a stack of modules
\begin{align}
    F_\vtheta = \bert_\vtheta \triangleq P_{\vtheta_{L+1}} \circ B_{\vtheta_L} \circ \cdots \circ B_{\vtheta_1} \circ E_{\vtheta_0} \\ 
    (\vtheta \triangleq \left[ \vtheta_l \right]_0^{L+1}), \nonumber 
\end{align}
among which $E$ is the embedding layer, $P$ a final pooling layer and each $B$ is a transformer block
\begin{align}
    B_\vvartheta: \vx \mapsto \lnorm(\vx + \dout(\mW_O\gelu(\mW_I\lnorm(\vx  \nonumber \\
    +\dout(\mW_D\attn(\mW_Q\vx, \mW_K\vx, \mW_V\vx)))))), 
\end{align}
where $\vvartheta \triangleq \left[ \mW_Q, \mW_K, \mW_V, \mW_D, \mW_I, \mW_O \right]$ collects all the learnable parameter matrices in the block. $\mW_Q$, $\mW_K$, $\mW_V$ and $\mW_D$ are the query, key, value, and dense self-attention projection matrices, respectively. $\mW_I$ and $\mW_O$ are the intermediate and output feedforward matrices, respectively. $\attn(\cdot, \cdot, \cdot)$ represents multi-head scaled dot-product attention~\citep{Vaswani2017}, $\dout(\cdot)$ dropout, $\lnorm(\cdot)$ layer normalization, and $\gelu(\cdot)$ the Gaussian error linear unit activation function~\citep{Hendrycks2016}.  
We experimented with the $\bert_\text{BASE}$ model~\citep{Devlin2018}, for which $L=12$, with total parameter count of 109M\,\footnote{
    Pre-trained parameters obtained from \url{https://github.com/google-research/bert}.
}.  
See Table~\ref{tab:param} for additional task-specific parameter counts, all 5 orders of magnitude smaller than the total parameter count. 
Optimization of them alone fails to fine-tune (see Appendix~\ref{app:last-layer-only}).

\subsection{GLUE benchmark}

The GLUE benchmark is a collection of diverse natural language understanding tasks~\citep{Wang2018}: CoLA~\citep{Warstadt2018}, SST~\citep{Socher2013}, MRPC~\citep{Dolan2005}, STS~\citep{Cer2018}, QQP~\citep{ShankarIyer2017}, MNLI~\citep{Williams2018}, QNLI~\citep{Rajpurkar2016}, and RTE~\citep{Dagan2006,Bar-Haim2006,Giampiccolo2007,Bentivogli2009}. 
We exclude the problematic WNLI set\,\footnote{
See (12) in \url{https://gluebenchmark.com/faq}.
}~\citep{Levesque2012}.  
We fine-tune on these tasks and report the evaluation performances. 
F1 is reported for QQP and MRPC, Matthews correlation for CoLA, Pearson and Spearman correlation for STS-B, and accuracy for all other tasks.

\subsection{Constrained fine-tuning procedures}

For all fine-tuning procedures, we use the exact hyperparameters as described in the original paper~\citep{Devlin2018} unless specified otherwise, with additional constraints described as follows. No constraints are imposed on task-specific last layers $G^{(t)}_\vphi$.

\paragraph{$L_0$-close fine-tuning}
To search for fine-tuned solutions that are $L_0$-close to the pre-trained parameters, we selectively fix the least sensitive parameter matrices at their pre-trained values and perform fine-tuning optimization on a lower-dimensional parameter space.  

\paragraph{Sparsification as fine-tuning}
In order to search for fine-tuned networks that are both sparse and $L_0$-close to the pre-trained one, we reparameterize the model by a multiplicative binary mask 
\begin{align}
    \vtheta = \tilde \vtheta \odot \vmu ,
\end{align}
where $\tilde \vtheta$ is the pre-trained parameters, and $\vmu \in \{0, 1\}^N$ the mask, $N$ being the dimensionality of the parameter space, and $\odot$ the Hadamard product.

If learning is purely through optimizing the mask $\vmu$ while holding $\tilde \vtheta$ constant, the mask is called a \emph{supermask}~\citep{Zhou2019}.  
Since $\vmu$ is discrete-valued and thus not differentiable, we reparameterize $\vmu$ as
\begin{align}
    \vmu = \bern(\sigmoid(\vnu)),
\end{align}
where $\bern(p)$ denotes an element-wise independent Bernoulli sampler with probability $p$, and $\sigmoid(\cdot)$ the sigmoid function, applied element-wise on $\vnu \in \sR^N$, the continuous \emph{mask parameter} that is task-specific.  
We treat gradient backpropagation through $\vmu$ as a straight-through estimator, similar to the techniques used in \cite{Mallya2018,Zhou2019}.
Same fine-tuning hyperparameters as described in \citet{Devlin2018} were used except for the learning rate (see Appendix~\ref{app:lr}).

Control over the final sparsity\,\footnote{
    Defined as the fraction of zero components, equal to one minus density.
} is exerted by initialization of $\vmu$ for fine-tuning.  
We initialize $\vnu$ according to a soft magnitude-based pruning mask: a fraction of small-magnitude values are initialized to $\nu=-5$ and the rest to $\nu = 5$. 
We found that the initial sparsity directly controls the final sparsity (see Appendix~\ref{app:corr-spar}), allowing us to produce masks with sparsity levels ranging from $1\%$ to $89\%$.

\section{Experimental results}
\label{sec:res}

\begin{table*}[t]
\caption{
    Distance metrics between fine-tuned and pre-trained parameters, compared against the expected values between two independent random initializations, either uniformly or normally distributed from $-\frac 1 {\sqrt{H}}$ to $\frac 1 {\sqrt{H}}$ where $H=768$ is the hidden dimension, as well as those between the pre-trained and a random initialization. 
    Statistics presented in the rightmost column are across all GLUE tasks.  
}
\label{tab:dist}
\vspace{-3mm}
\begin{center}
\begin{small}
\setlength\tabcolsep{3.6pt}
\begin{tabular}{l l l l l l}
    \toprule
    Distance metric & 
    \begin{tabular}{@{}l@{}}
        Between uniform \\ initializations 
    \end{tabular} & 
    \begin{tabular}{@{}l@{}}
        Between normal \\ initializations 
    \end{tabular} & 
    \begin{tabular}{@{}l@{}}
        Between uniform \\ initialization \\ and pre-trained 
    \end{tabular} & 
    \begin{tabular}{@{}l@{}}
        Between normal \\ initialization \\ and pre-trained  
    \end{tabular} & 
    \begin{tabular}{@{}l@{}}
        Between fine-tuned \\ and pre-trained \\ $([\min, \max])$ 
    \end{tabular} \\
    \midrule
    $L_1$-distance & $20.0 \pm 0.1$ & $16.7 \pm 0.1$ & $41.9 \pm 15.2$ & $40.9 \pm 15.4$ & $[0.1, 3.3]$ \\
    Angular distance & $0.500$ & $0.500$ & $0.500$ & $0.500$ & $[0.001, 0.027]$ \\
    \bottomrule
\end{tabular}
\end{small}
\end{center}
\vskip -0.1in
\end{table*}

\begin{figure}[t]
  \centering
  \begin{subfigure}[b]{0.40\textwidth}
    \phantomsubcaption
    \includegraphics[width=\textwidth]{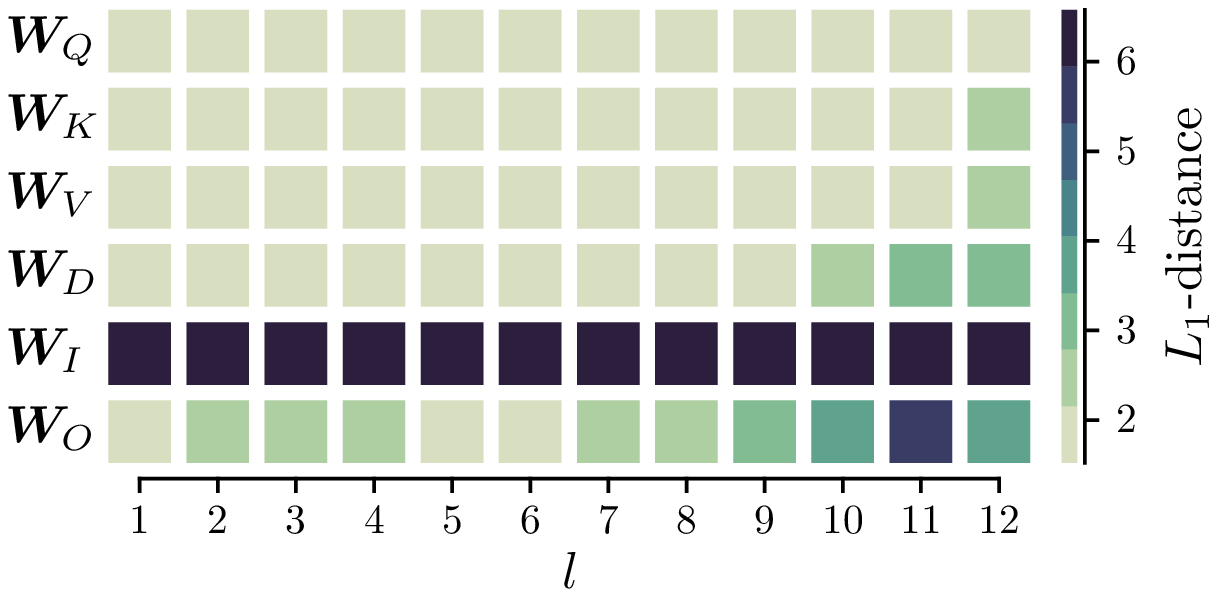} 
    \label{fig:layer-closeness-l1}
  \end{subfigure}
  \quad
  \begin{subfigure}[b]{0.40\textwidth}
    \phantomsubcaption
    \includegraphics[width=\textwidth]{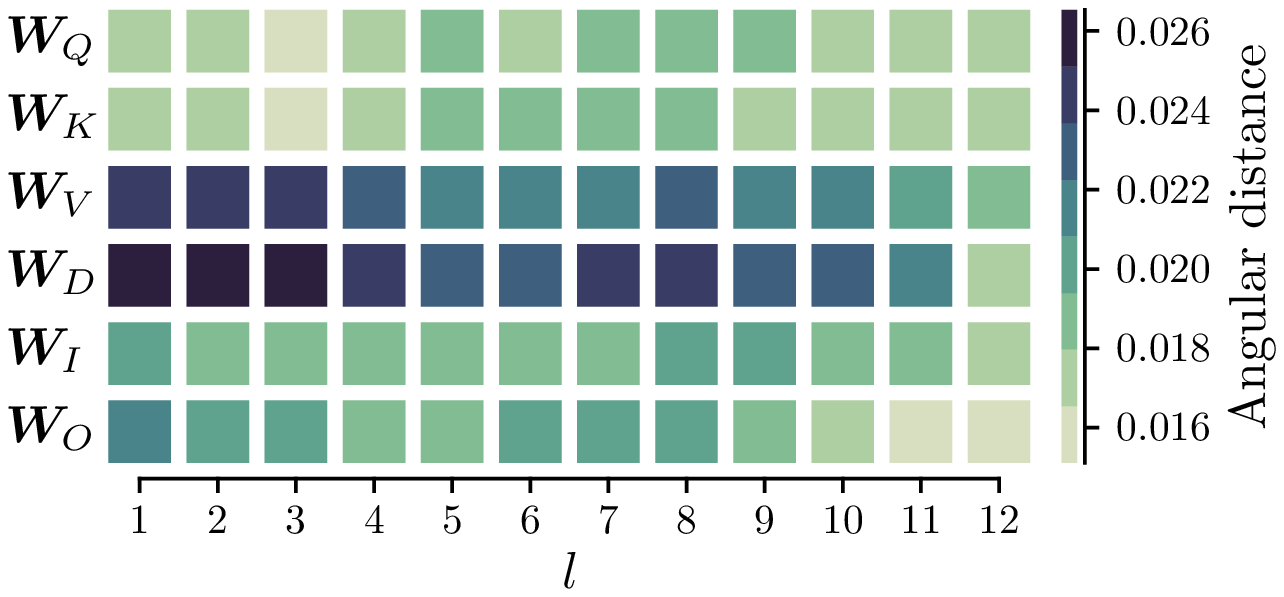} 
    \label{fig:layer-closeness-ang}
  \end{subfigure}
  \vspace{-8pt}
  \caption{
    $L_1$- and angular distances in parameter subspaces between pre-trained and fine-tuned weights.
	Shown are metrics across the 12 encoder stack layers for the self-attention projection matrices ($\mW_Q$, $\mW_K$, $\mW_V$ and $\mW_D$) and feed-forward matrices ($\mW_I$ and $\mW_O$). 
	The results presented here are for MNLI fine-tuning, but similar patterns are observed across all GLUE tasks. 
}
\label{fig:layer-closeness}
\end{figure}

\begin{table*}[t]
\caption{
    $L_0$-close fine-tuning results: layers excluded from fine-tuning, corresponding number of parameters remaining to fine-tune, and the fine-tuning performance on the MRPC task (F1 score); other GLUE tasks show similar patterns. 
    We report the mean and standard deviation across 10 independent runs. 
}
\label{tab:l0-close}
\vspace{-3mm}
\begin{center}
\begin{small}
\setlength\tabcolsep{3.6pt}
\begin{tabular}{l c c}
    \toprule
     Layers excluded from fine-tuning & Task-specific parameter storage & F1 score \\ 
     \midrule
     None (baseline) & 109M float ($100\%$) & $89.4 \pm 0.7$ \\
     (1) Key projection layers in self-attention & 102M float ($94\%$) & $89.1 \pm 0.8$ \\
     (2) Deepest encoder stack layers & 95M float ($87\%$) & $88.8 \pm 0.9$ \\
     (3) Word embedding layer & 86M float ($78\%$) & $89.3 \pm 0.8$ \\
     (1), (2), and (3) & 66M float ($60\%$) & $88.7 \pm 0.9$ \\
     \midrule
     (1), (2), and (3) with $30\%$ sparse fine-tuning  & 66M binary ($60\%$) & $87.4 \pm 2.2$ \\
     (1), (2), and (3) with $40\%$ sparse fine-tuning  & 66M binary ($60\%$) & $86.6 \pm 2.2$ \\
    \bottomrule
\end{tabular}
\end{small}
\end{center}  
\vskip -0.1in
\end{table*}

\subsection{Fine-tuned and pre-trained parameters are $L_1$-close and angular-close}

We observe that the original fine-tuning procedures for GLUE tasks all take $10^2$ to $10^4$ parameter update steps (Table~\ref{tab:param}), negligible compared to the dimensionality of the parameter space, \textit{viz.} $10^8$.
Thus, we first asked whether fine-tuned parameters are indeed close to the pre-trained ones in parameter space.  
We measured the $L_1$-distances, \textit{i.e.} $L_1$-norm of parameter difference, and angular distances (Table~\ref{tab:dist}).
Specifically, we inspect the weight matrices in all self-attention layers, of size $768\times768$ where $768$ is the hidden state dimension. 
We report the minimum and maximum values across GLUE tasks: RTE showed the smallest values, and MNLI showed the largest values. 
Evidently, we see a significantly higher $L_1$- and angular-closeness between fine-tuned and pre-trained parameters as compared to 
the expected distance between two independent random initializations, or that between an initialization and the pre-trained paremeters.  
This confirms that, during the course of fine-tuning, the very few model parameter updates traversed a very short distance in the parameter space. Comparing the parameter distance across GLUE tasks, we find that it scales with the number of fine-tuning iterations (see Appendix~\ref{app:corr-dist-step}).

Further, we inspect the closeness in parameter subspaces for each layer.  
We found that, though all layers change very little during fine-tuning, there is nevertheless a high degree of variability across different parameter matrices (Figure~\ref{fig:layer-closeness}). 
Blocks deeper in the encoder stack are less $L_1$-close but more angular-close than shallower ones.
In all self-attention modules, value and dense projection matrices ($\mW_V$ and $\mW_D$) change considerably more than query and key projection matrices ($\mW_Q$ and $\mW_K$) during fine-tuning.  

\subsection{$L_0$-close fine-tuning}

Inspired by the high degree of variability in each layer's parameter change during fine-tuning, we ask whether effective fine-tuning can be achieved by optimizing only a fraction of layers while having others fixed at pre-trained values, resulting in fine-tuned models $L_0$-close in parameter space.  

Our results suggest this is indeed feasible (Table~\ref{tab:l0-close}).
Informed by different layers' sensitivity to fine-tuning, we performed fine-tuning experiments by progressively excluding: (1) key projection layers in self-attention across all encoder stack layers, (2) the penultimate and ultimate encoder stacks, and (3) the word embedding layer.
Each of these exclusions independently or all three combined do not substantially degrade performance, while reducing the number of parameters to fine-tune by up to $40\%$ (from 109M to 66M).

\begin{figure}[t]
    \centering
    \includegraphics[width=0.40\textwidth]{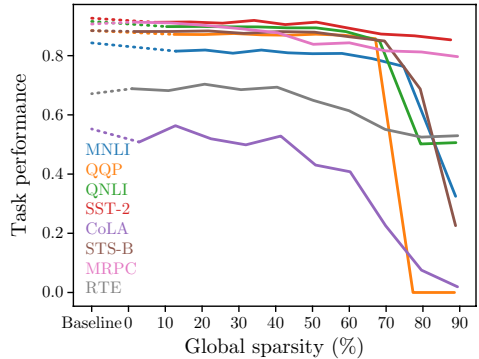}
    \caption{
        Performance of supermask fine-tuned models across GLUE tasks. 
        We show the mean of performance metrics across 10 independent Bernoulli sampling procedures. 
        Note the baseline performance for each task marked by the leftmost end of each curve. 
    }
    \label{fig:supermask_perf}
\end{figure}

\begin{figure}[t]
    \centering
    \includegraphics[width=0.40\textwidth]{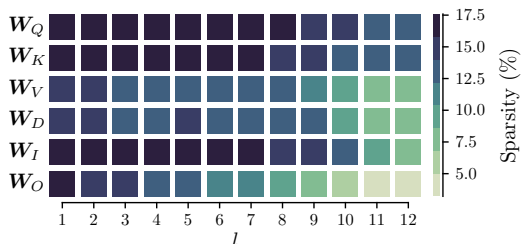}
    \caption{
    Supermask sparsity levels across layers. Shown is the low-sparsity MNLI supermask with a global sparsity level of $12.9\%$; similar patterns are observed across all GLUE tasks.
    }
    \label{fig:sparsity_layers}
\end{figure}

\subsection{Sparsification as fine-tuning}

Encouraged by these results, we ask whether more aggressive constraints can be imposed on the fine-tuning process to further reduce computational cost.
Though $L_0$-close fine-tuning obviates optimization of a substantial fraction of parameters, avoiding full storage of all parameters for each fine-tuned task, all operations still need to be performed at inference time.
In order to reduce operations, we seek to sparsify parameters.  
This amounts to a search over a binary mask in a high-dimensional parameter space.  
We adopt \emph{supermask} training (see Section~\ref{sec:meth}) to this end.

Figure~\ref{fig:supermask_perf} shows fine-tuned model performance across GLUE tasks obtained by supermask training.  
Final sparsity level of the supermask is controlled by its initialization (see Section~\ref{sec:meth} and Appendix~\ref{app:corr-spar}). 
We note that there is little task performance degradation between $1\%$ and $40\%$ parameter sparsity, very close to sparse networks produced by iterative pruning~\citep{Zhu2017} but underperfoming it at high sparsity levels (see Appendix~\ref{app:sm-vs-p}).
Layer-wise sparsity levels of supermasks also demonstrate systematic trends (Figure~\ref{fig:sparsity_layers}): (1) across GLUE tasks, $\mW_Q$, $\mW_K$ and $\mW_I$ tend to be sparser than $\mW_V$, $\mW_D$ and $\mW_O$, and (2) shallower encoder stack layers are sparser than deeper ones.  
Moreover, we show that supermask fine-tuning of only a fraction of sensitive layers could also achieve performance with little degradation from baseline (Table~\ref{tab:l0-close}).  

\subsection{Many good, sparse fine-tuned supermasks exist, but for pre-trained parameters only}

One surprising finding of this study is the many occurrences of \emph{good} fine-tuned parameters among the $2^N$ configurations in the set $\left\{ \vtheta: \vtheta = \tilde \vtheta \odot \vmu \: \middle| \: \vmu \in \{0, 1 \}^N \right\}$ \textit{viz.} vertices of an $N$-dimensional hypercube, even though most of them are quite distant from the pre-trained parameters by $L_1$-metric. 

First, there exist supermasks up to $40\%$ sparse without remarkable performance degradation for all GLUE tasks, for some tasks even sparser (Figure~\ref{fig:supermask_perf}, right end).
Second, for any task, below this maximum sparsity, we found good masks at any sparsity level (Figure~\ref{fig:supermask_perf}), which can be controlled by initialization of the supermask (see Appendix~\ref{app:corr-spar}).
Finally, while it is natural that performance drops as the mask becomes extremely sparse (Figure~\ref{fig:supermask_perf}, right end), it is rather counterintuitive that there exist good supermasks at the dense extreme (Figure~\ref{fig:supermask_perf}, left end), since we observe that the pre-trained model with only the task-specific last layer fine-tuned utterly fails to learn any task (Appendix~\ref{app:last-layer-only}).
Noticeably, good supermasks selectively prune important weights of large magnitudes (Appendix~\ref{app:sm-non-trivial}).

\begin{figure}[t]
    \centering
    \includegraphics[width=0.40\textwidth]{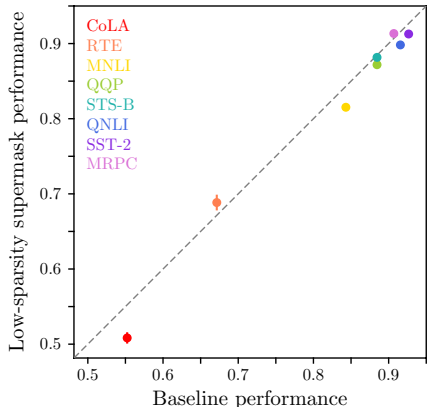}
    \caption{
    Low-sparsity supermask performance, \textit{i.e.} task performance of super-masks initialized at $0\%$ sparsity, compared against baseline.
    }
    \label{fig:low_sparsity}
\end{figure}

\begin{table*}[t]
\caption{
    Low-sparsity supermask performance. 
    We report the sparsity levels achieved when the supermasks were initialized at $0\%$ sparsity. 
    For several tasks, fine-tuning is achieved with less than $3\%$ of pre-trained weights pruned. 
    For the supermask evaluation results, we include the mean and standard deviation of 10 Bernoulli samplings of a single run. 
}
\label{tab:low_sparsity}
\vspace{-3mm}
\begin{center}
\begin{small}
\setlength\tabcolsep{3.6pt}
\begin{tabular}{l c c c c c c c c}
    \toprule
    {GLUE Task} & MNLI & QQP & QNLI & SST-2 & CoLA & STS-B & MRPC & RTE \\ 
    \midrule
    \multirow{1}{*}{\begin{tabular}{@{}l@{}}Baseline\end{tabular}} & $84.3/85.6$ & $88.5$ & $91.6$ & $92.7$ &  $55.2$ & $88.5$ & $90.7$ & $67.1$ \\
    \multirow{2}{*}{\begin{tabular}{@{}l@{}}Supermask\end{tabular}}
    & $81.5/82.9$ & $87.2$ & $89.8$ & $91.3$ & $50.8$ & $88.2$ & $91.3$ & $68.8$ \\
    & $\pm 0.1$ & $\pm 0.1$ & $\pm 0.1$ & $\pm 0.2$ & $\pm 0.8$ & $\pm 0.1$ & $\pm 0.4$ & $\pm 1.0$ \\
    \midrule
    {Final sparsity} & $12.9\%$ & $12.6\%$ & $10.3\%$ &  $7.4\%$ & $2.9\%$ & $2.2\%$ & $1.3\%$ & $1.0\%$ \\
    \bottomrule
\end{tabular}
\end{small}
\end{center}  
\vskip -0.1in
\end{table*}

\begin{figure}[t]
	\centering
	\begin{subfigure}[b]{0.40\textwidth} 
		\includegraphics[width=\textwidth]{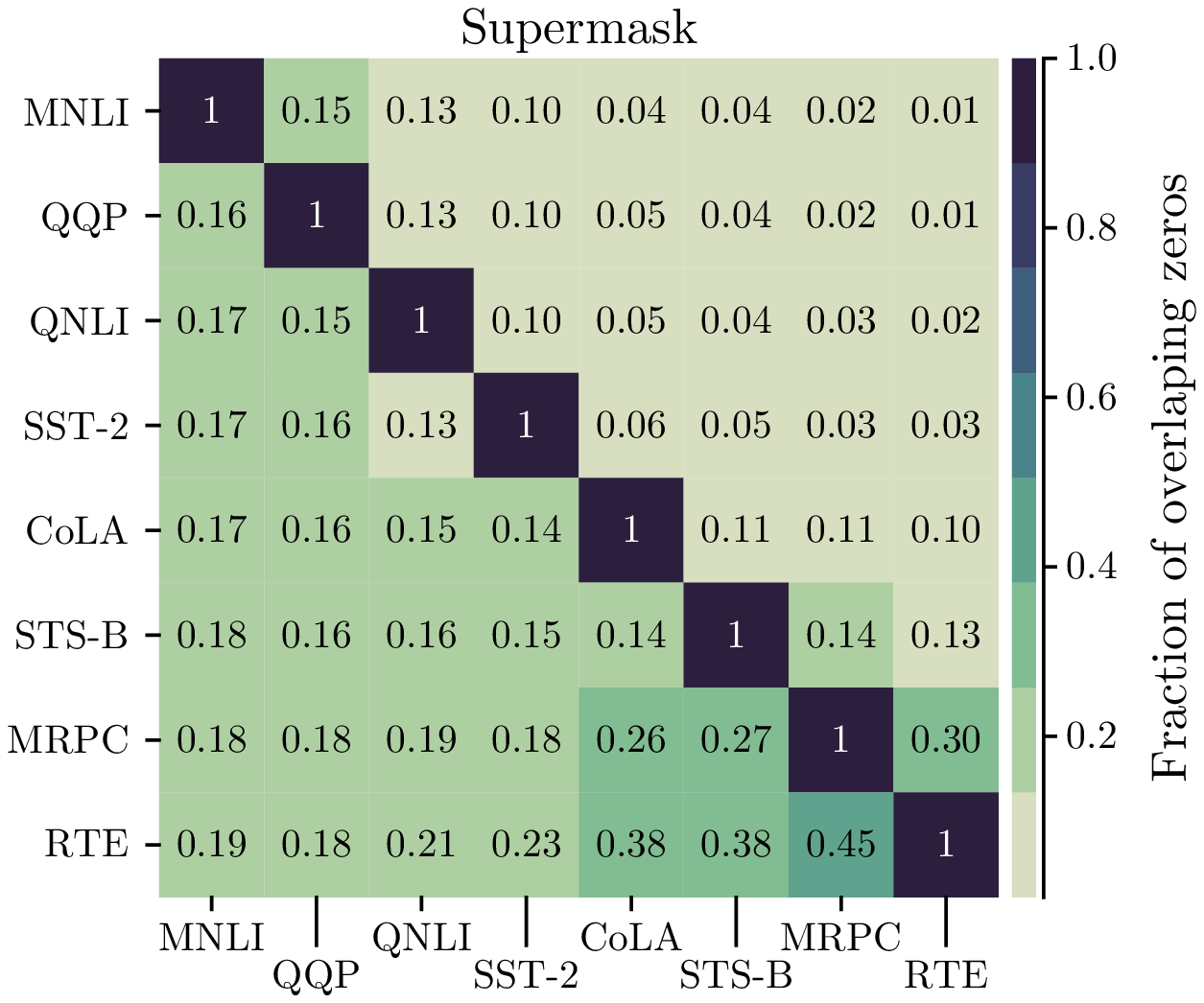}
	\end{subfigure}
	\vspace{1em} 
	\begin{subfigure}[b]{0.40\textwidth} 
		\includegraphics[width=\textwidth]{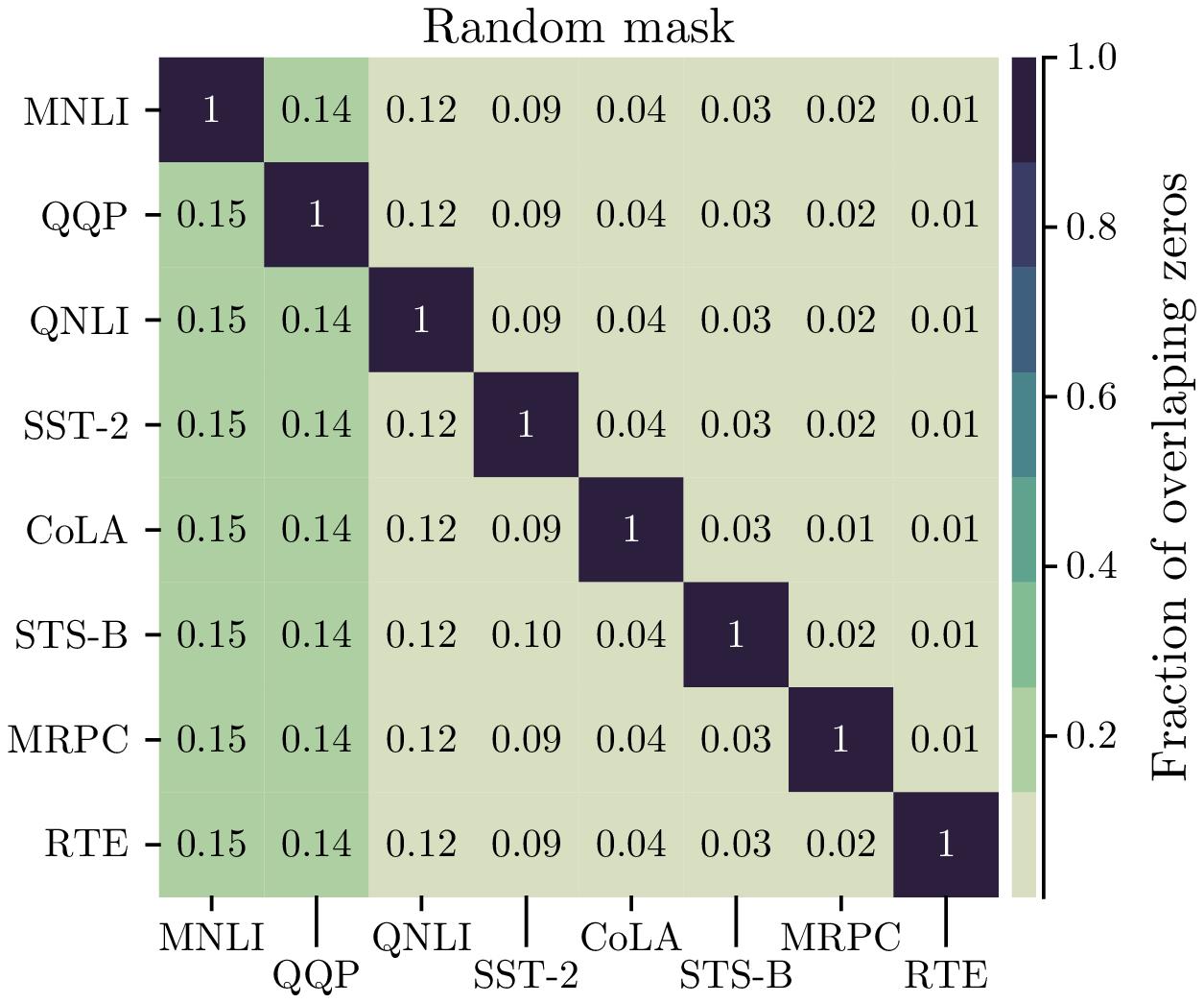}
	\end{subfigure}
	\caption{
	Fractions of overlap of zero elements in supermasks across GLUE tasks, compared to randomly generated masks. 
	Each value in the grid shows the fraction of pruned elements in one task (horizontal axis) that are also pruned in the other (vertical axis). 
	Here, we show low-sparsity supermasks (initialized at $0\%$ sparsity) and compare the masks in the value layer of the first encoder, which is one of the most sparse layers in the entire model.
	}
	\label{fig:overlap}
\end{figure}

To understand this phenomenon better, we study the supermasks trained with all-dense initialization (Figure~\ref{fig:low_sparsity}).
Surprisingly, these low-sparsity supermasks successfully learn to perform all the tasks without noticeable degradation from baseline.  
Essentially, complicated tasks like MNLI and QQP can be learned by clamping $12$-$13\%$ of the pre-trained weights to zero (see Appendix~\ref{app:supermask-learning-curve} for how model performance improves with sparsity), whereas for simple tasks like MRPC and RTE, setting only $1$-$2\%$ of the pre-trained weight entries to zero suffices to learn the task (Table~\ref{tab:low_sparsity}). 
Fine-tuning can indeed be very \emph{fine}, suggesting relatively frequent occurrences of \emph{good} solutions within a sparse $L_0$-neighborhood of the pre-trained parameters. 

Moreover, we ask whether such frequent occurrences of good sparsified versions of parameters is a unique property of the pre-trained weights.  
In other words, can one also obtain good supermasks on parameters that are not pre-trained?  
To answer this question, we perform supermask fine-tuning on pre-trained parameters with components shuffled (thusly norm-preserved). 
Performance degrades significantly, for instance, for the MRPC task, from an F1 score of $91.3$ to $81.2$ with shuffled pre-trained parameters. 
It is clear that one cannot obtain any good masks by doing so, suggesting that having high-performance sparsified versions is unique to pre-trained parameters.

\subsection{Task-uniqueness of fine-tuned supermasks}

Finally, we ask whether the supermasks learned to perform different tasks share commonalities.  
Specifically, we quantify the amount of overlapping zeros in learned supermasks across different tasks (Figure~\ref{fig:overlap}). 
It seems the overlaps are not substantially larger than what would have been caused by chance, suggesting that, even though there seem to be many good supermasks for each task, these masks are largely distinct from each other, each unique to the task it learns.

\section{Discussion}
\label{sec:disc}

One very puzzling fact about modern deep neural networks is that overparameterization helps both generalization and optimization.  
On the one hand, given an effective network architecture reflecting proper inductive biases, better generalizing models are always larger models~\citep{Hestness2017}.  
On the other hand, sheer increases in dimensionality of the parameter space seldom make stochastic gradient-based optimization more difficult: deeper and/or wider networks take just about the same, if not a lower number of training iterations to converge.  
For example, ResNet-18 (11.7M parameters) and ResNet-152 (60.2M parameters) both train to converge, at similar convergence rates, in no more than 600K iterations on Imagenet~\citep{He2015}.  
Thus, given adequate computing infrastructure, one always trains the largest possible model in order to obtain the best performance.  
This is perhaps most prominent in recent pre-trained huge language models~\citep[\textit{e.g.}][]{Devlin2018,Radford2018, Shoeybi2019} that have achieved state-of-the-art performance on language comprehension tasks.  
Similarly, fine-tuning larger pre-trained language models is just as easy, if not easier, than fine-tuning smaller ones.  
Fine-tuning steps are usually five orders of magnitude smaller than the dimensionality of the parameter space (Table~\ref{tab:param}).  
A direct consequence of this is that, in the parameter space, fine-tuned networks do not deviate substantially from the pre-trained one, which we quantitatively establish in this study.  
Analogous to the contrast between the low generalization performance of small networks and the high compressibility of large networks in the case of ResNets~\citep[\textit{e.g.}][]{Zhu2017,Frankle2018}, we are faced with the high generalization performance of large language models and the low level of dissimilarity before and after fine-tuning.  
Just as network compression can generate compact models for efficient inference, the abovementioned parameter closeness can also be taken advantage of to achieve efficient computation, which we demonstrate in this work.

We show that, due to surprisingly frequent occurrences of good parameter configurations in a close $L_0$-neighborhood and in the set of sparsified large pre-trained language models, two techniques are highly effective in producing efficient fine-tuned networks to perform specific language understanding tasks: (1) optimizing only the most sensitive layers and (2) learning to sparsify pre-trained parameters as fine-tuning. 
In contrast to commonly employed \emph{post-training} sparsification methods which always incur performance degradation, our procedure of sparsifying pre-trained networks \citep[similar to][]{Mallya2018,Mancini2019} is by itself an optimization process that learns specific tasks.

\section{Acknowledgements}
We thank Sofia Samaniego de la Fuente for help with the experiments. We also wish to thank Robert S. Schreiber, Urs K{\"o}ster, Jorge Albericio, Natalia S. Vassilieva, and Marcel Nassar for discussions and feedback on the manuscript. 

\bibliography{references}

\clearpage
\begin{appendices} 
\appendix

\section{Optimization of task-specific last layers alone fails to fine-tune}
\label{app:last-layer-only}

Optimization of only task-specific layers does not lead to successful fine-tuning. 
For instance, for the MRPC task, freezing parameter weights in the pre-trained model and optimizing the task-specific last layer alone yields a non-performing model.  
Across 10 independent runs, the model consistently predicts all $1$'s for the paraphrase classification task, yielding an F1 score of $81.2$. 
This is a significant degradation compared to the baseline performance of $89.4 \pm 0.7$ across multiple runs (Table~\ref{tab:l0-close}). 
Thus, it is critical to fine-tune layers in the pre-trained model and not just the task-specific layers alone.

\section{Learning rate of supermask training}
\label{app:lr}

Supermask training requires a much larger learning rate compared to typical training~\citep{Zhang2019}. 
While a learning rate of $2\times10^{-5}$ is used for optimizing weights, a learning rate of $2\times10^{-1}$ is used for optimizing masks. 
We notice a degradation in performance at smaller learning rates for supermask training (Table~\ref{tab:learning_rate}).
This pattern holds true across GLUE tasks.

\begin{table}[h]
\caption{
    MRPC low-sparsity supermask performance at learning rates from $2\times10^{-5}$ and $2\times10^{-1}$. 
}
\label{tab:learning_rate}
\vspace{-3mm}
\begin{center}
\begin{small}
\setlength\tabcolsep{3.6pt}
\begin{tabular}{c c}
    \toprule
    {Learning-rate} & \multirow{1}{*}{\begin{tabular}{@{}l@{}}F1 score\end{tabular}} \\
    \midrule
    $2\times10^{-1}$ & $91.3 \pm 0.4$\\
    $2\times10^{-2}$ & $82.0 \pm 0.2$\\
    $2\times10^{-3}$ & $0.0$\\
    $2\times10^{-4}$ & $0.0$\\
    $2\times10^{-5}$ & $0.0$\\
    \bottomrule
\end{tabular}
\end{small}
\end{center}  
\vskip -0.1in
\end{table}

\section{Correlation between initial and final sparsities of supermasks}
\label{app:corr-spar}

\begin{figure}[h]
    \centering
    \includegraphics[width=0.40\textwidth]{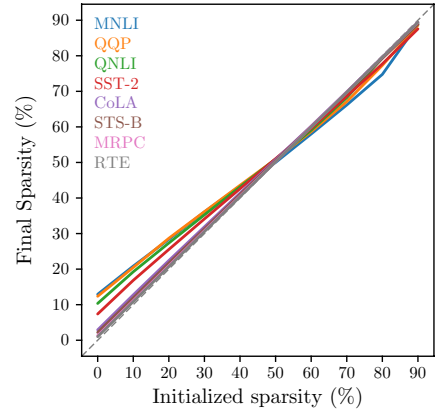}
    \caption{
    Initial versus final sparsity levels of supermasks. 
    }
    \label{fig:init_final_sparsity}
\end{figure}

There is no straightforward control of the amount of weights pruned in previous reports of supermask training~\citep{Zhang2019,Mallya2018}. 
We find that setting the initial sparsity through a soft magnitude-based pruning mask controls the final sparsity level, which we use to produce supermasks of varied sparsity levels. 
Figure~\ref{fig:init_final_sparsity} shows this correlation between initial and final sparsities of supermasks for different GLUE tasks.
We note that, at lower initial sparsity levels, the supermask is pushed to a greater sparsity level, whereas at higher sparsity levels, the supermask is pushed to a lower sparsity level. 
This pattern is similar across GLUE tasks but is most prominent in the MNLI task, scaling with the number of fine-tuning steps (Table~\ref{tab:task_info}).

\section{Correlation of parameter distance with fine-tuning steps}
\label{app:corr-dist-step}

\begin{figure}[h]
	\centering
	\includegraphics[width=0.40\textwidth]{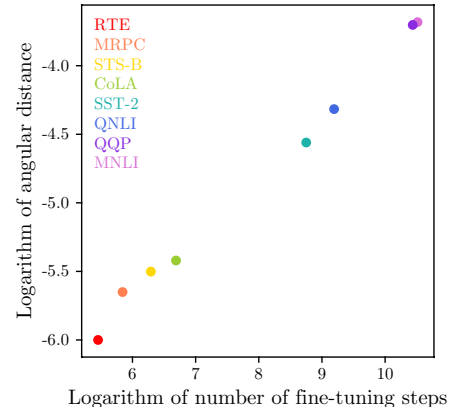}
	\caption{
	Correlation of parameter distance with the number of fine-tuning iterations. 
	Shown are angular distances.  
	Each data point corresponds to a different GLUE task.
	}
	\label{fig:param_dist}
\end{figure}

In order to understand how distance in parameter space increases as a function of fine-tuning steps, we study this relationship across GLUE tasks.  
We find that parameter distance scales with the number of fine-tuning steps by a power law with exponent close to $0.5$ (Figure \ref{fig:param_dist}).

\section{Fine-tuning with iterative pruning}
\label{app:sm-vs-p}

We also use iterative pruning~\citep{Zhu2017} during fine-tuning to produce sparse models.
Pruning is based on weight magnitudes in each layer and is performed periodically during fine-tuning with sparsity gradually increasing from $0\%$ to a final level according to a cubic schedule. 

\begin{figure}[h]
    \centering
    \includegraphics[width=0.40\textwidth]{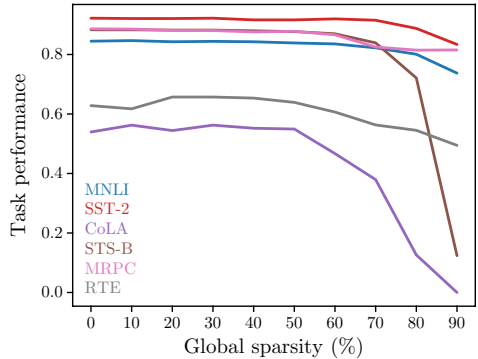}
    \caption{Iterative pruning during fine-tuning. We plot the evaluation performance at sparsity levels from $10\%$ to $90\%$ across GLUE tasks. Note the baseline performance for each task marked by the leftmost end of each curve ($0\%$ sparsity).}
    \label{fig:itr_pruning}
\end{figure}

Iterative pruning during fine-tuning (Figure~\ref{fig:itr_pruning}) outperforms supermask training (Figure~\ref{fig:supermask_perf}) at higher sparsity levels. 
While supermask training remains successful up to $40\%$ sparsity, iterative pruning produces binary masks up to $50\%$ sparse and for some tasks even sparser without significant performance degradation. 
Though iterative pruning produces sparse models, the fine-tuned models do not share parameters--one still needs to store all parameters for each task. 
Fine-tuned supermasks, on the other hand, store only a binary mask of certain layers for each task, with all tasks sharing a same set of underlying pre-trained weights.

\begin{figure}[t]
  \centering
  \begin{subfigure}[b]{0.40\textwidth}
    \phantomsubcaption
    \includegraphics[width=\textwidth]{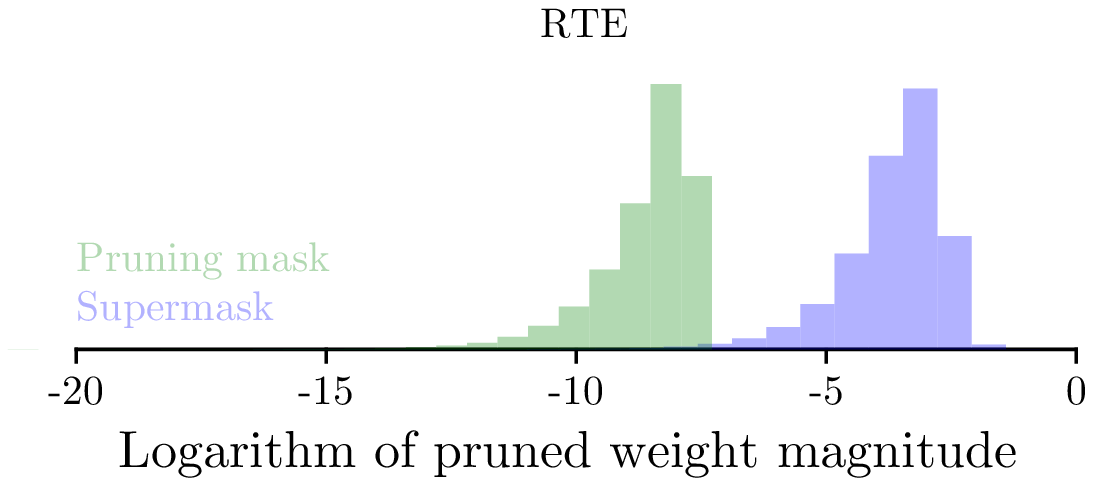} 
  \end{subfigure}
  \quad
  \begin{subfigure}[b]{0.40\textwidth}
    \phantomsubcaption
    \includegraphics[width=\textwidth]{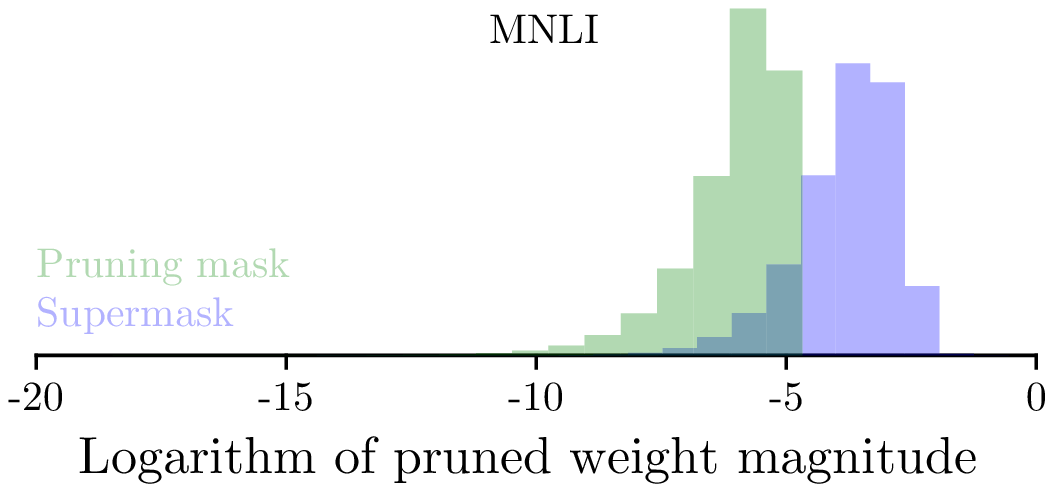} 
  \end{subfigure}
	\caption{
	Pruned weight distributions, compared between supermask and magnitude-based pruning. Shown for the RTE and MNLI fine-tuning tasks.
	}
	\label{fig:sm-nontrivial}
\end{figure}

\section{Fine-tuned supermasks are not trivial}
\label{app:sm-non-trivial}

How does the learning of a supermask actually work?  
Does a supermask simply learn to prune away the weights with smallest magnitudes? 
Since pure magnitude-based pruning of pre-trained weights does not perform any task-specific learning, we reason that the weight entries being set to zero by the supermask must have significant values. 
Here, we inspect the magnitudes of the pre-trained weights zeroed by the supermasks (Figure~\ref{fig:sm-nontrivial}, Table~\ref{tab:sm-vs-prune}).
These weights turn out to have remarkably higher magnitudes than the smallest entries, suggesting the learning of supermasks is not trivial magnitude-based pruning. 

\begin{table*}[t]
\caption{
    Comparison between weights pruned with low-sparsity supermasks (initialized at $0\%$ sparsity) and weights pruned with magnitude-based pruning at the same final sparsity. 
    We report the maximum and mean magnitude of the pruned weights. 
    The last row shows percentages of the overlap between the supermask and the magnitude-based pruning mask, \textit{i.e.} the percentages of weights zeroed by the supermask that are also the smallest weights. 
}
\label{tab:sm-vs-prune}
\vspace{-3mm}
\begin{center}
\begin{small}
\setlength\tabcolsep{3.6pt}
\begin{tabular}{l c c c c c c c c}
    \toprule
    GLUE task & MNLI & QQP & QNLI & SST-2 & CoLA & STS-B & MRPC & RTE \\ \midrule
    Pruned max          & $0.0093$ & $0.0093$ & $0.0075$ & $0.0059$ & $0.0022$ & $0.0018$ & $0.0009$ & $0.0007$ \\
    Supermask max       & $1.7$ & $6.4$ & $2.5$ & $1.7$ & $1.1$ & $2.8$ & $1.8$ & $2.8$ \\ \midrule 
    Pruned mean         & $0.0033$ & $0.0032$ & $0.0026$ & $0.0020$ & $0.0008$ & $0.0006$ & $0.0003$ & $0.0002$ \\
    Supermask mean      & $0.032$ & $0.033$ & $0.033$ & $0.035$ & $0.037$ & $0.036$ & $0.038$ & $0.036$ \\ \midrule
    Overlap             & $11.1\%$ & $10.0\%$ & $6.7\%$ & $3.6\%$ & $0.7\%$ & $0.7\%$ & $0.7\%$ & $0.7\%$ \\
    \bottomrule
\end{tabular}
\end{small}
\end{center}  
\vskip -0.1in
\end{table*}

\section{Learning curves of low-sparsity supermask fine-tuning}
\label{app:supermask-learning-curve}

\begin{figure}[h]
    \centering
    \includegraphics[width=0.40\textwidth]{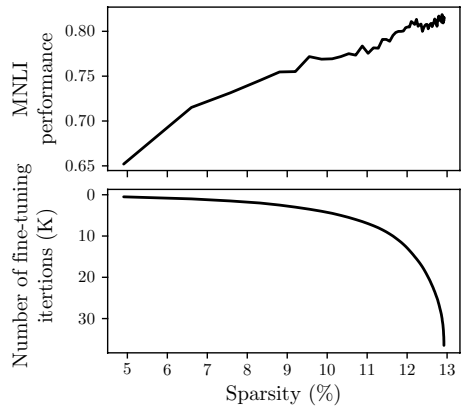}
    \caption{
    Learning curves of MNLI low-sparsity supermask fine-tuning. 
    }
    \label{fig:mnli-step-sp-perf}
\end{figure}

Our results suggest that supermask fine-tuning, if initialized at $0\%$ sparsity, gradually increases sparsity during optimization, reaching a final sparsity level that correlates with the number of fine-tuning steps (Table~\ref{tab:low_sparsity}).  
For MNLI, the GLUE task with the most fine-tuning steps, the sparsity level reaches $12.9\%$.  
We ask how prediction accuracy grows with sparsity during fine-tuning.  
As shown in Figure~\ref{fig:mnli-step-sp-perf}, like model performance, sparsity rapidly grows during the initial phase of fine-tuning.  
This makes model performance increase roughly linearly with sparsity.   

\end{appendices} 

\end{document}